\documentclass[10pt,twocolumn,letterpaper]{article}

 \usepackage{cvpr}              % To produce the CAMERA-READY version
\usepackage{xcolor}

\makeatletter
\@ifundefined{red}{}{}
\@ifundefined{todo}{}{}
\makeatother

\usepackage{booktabs}
\usepackage{algorithm}
\usepackage{algorithmic}
\usepackage{amssymb}
\usepackage{amsmath}
\usepackage{multirow}
\usepackage{multicol}
\usepackage{xcolor}
\usepackage{graphicx}
\usepackage{float}
\usepackage{tcolorbox} 
\usepackage[table,xcdraw]{xcolor}
\usepackage{xcolor} 
\definecolor{myPink}{RGB}{255,226,237}
\definecolor{myTeal}{RGB}{0,161,106}
\definecolor{myOrange}{RGB}{249,70,0}
\definecolor{myBule}{RGB}{0,104,211}
\definecolor{reda}{RGB}{202,0,0}
\definecolor{mygreen}{RGB}{0,136,51}
\usepackage[table,xcdraw]{xcolor}
\usepackage{tikz}

\usepackage{xcolor}

\makeatletter
\@ifundefined{red}{}{}
\@ifundefined{todo}{}{}
\makeatother

\definecolor{cvprblue}{rgb}{0.21,0.49,0.74}
\definecolor{task1}{rgb}{0.5,0.49,0.74}
\definecolor{task2}{rgb}{0.21,0.5,0.74}
\definecolor{task3}{rgb}{0.21,0.49,0.5}
\usepackage[pagebackref,breaklinks,colorlinks,allcolors=cvprblue]{hyperref}

\def\paperID{11023} % *** Enter the Paper ID here
\def\confName{CVPR}
\def\confYear{2026}

\title{TP-Seg: Task-Prototype Framework for Unified Medical Lesion Segmentation}

\author{
Jiawei Xu$^{1}$ \quad
Qiangqiang Zhou$^{1}$\thanks{Corresponding authors. Qiangqiang Zhou (qiang@jxnu.edu.cn) and Xiaoqi Zhao (xiaoqi.zhao@yale.edu).} \quad
Dandan Zhu$^{2}$ \quad
Yong Chen$^{1}$ \quad
Yugen Yi$^{1}$ \quad
Xiaoqi Zhao$^{3*}$\\
$^{1}$School of Artificial Intelligence, Jiangxi Normal University\\
$^{2}$East China Normal University\\
$^{3}$Yale University
}

\begin{document}
\maketitle
\begin{abstract}
Building a unified model with a single set of parameters to efficiently handle diverse types of medical lesion segmentation has become a crucial objective for AI-assisted diagnosis. Existing unified segmentation approaches typically rely on shared encoders across heterogeneous tasks and modalities, which often leads to feature entanglement, gradient interference, and suboptimal lesion discrimination. In this work, we propose TP-Seg, a task-prototype framework for unified medical lesion segmentation. On one hand, the task-conditioned adapter effectively balances shared and task-specific representations through a dual-path expert structure, enabling adaptive feature extraction across diverse medical imaging modalities and lesion types. On the other hand, the prototype-guided task decoder introduces learnable task prototypes as semantic anchors and employs a cross-attention mechanism to achieve fine-grained modeling of task-specific foreground and background semantics. Without bells and whistles, TP-Seg consistently outperforms specialized, general and unified segmentation methods across 8 different medical lesion segmentation tasks covering multiple imaging modalities, demonstrating strong generalization, scalability and clinical applicability. Code and pretrained models will be available at the
\href{https://github.com/jiaweiXu1029/TP-Seg}{\textcolor{myOrange}{link}}
% Unified medical lesion segmentation aims to process multiple types of lesions using a single model and a unified set of training parameters. Although existing methods have shown good potential, their reliance on shared encoders often leads to feature overlap and gradient interference. Therefore, we propose the first task-conditioned adapter specifically designed for unified medical lesion segmentation tasks. This adapter effectively balances shared encoding and task-specific encoding during the feature extraction phase through a "shared and task-specific dual-path expert" structure. Furthermore, to overcome the limitations of conventional decoders in capturing foreground and background semantics, we design a Prototype-Guided Task Decoder that uses learnable task prototypes as semantic anchors and employs a cross-attention mechanism to achieve precise modeling of task-specific foreground and background semantics. Based on these designs, we construct a novel unified medical lesion segmentation framework, TP-Seg. Extensive experimental results demonstrate that TP-Seg significantly outperforms existing methods across 8 medical lesion segmentation tasks, greatly enhancing unified segmentation performance. More importantly, TP-Seg establishes a new paradigm for generalized multi-lesion evaluation, allowing a single model trained once to effectively segment diverse lesion types across different tasks and modalities, providing new insights and directions for the clinical application of medical lesion segmentation.
\end{abstract}
    
\section{Introduction}
\label{sec:intro}
Medical lesion segmentation (MLS) is a fundamental task in computer-aided diagnosis, playing a vital role in clinical decision-making, lesion localization, and treatment assessment. MLS encompasses a wide variety of lesion types, such as wet age-related macular degeneration~\cite{wetamd}, brain tumors~\cite{BTD1}, adenocarcinoma~\cite{ebhi}, thyroid nodules~\cite{TNUI}, colorectal polyps~\cite{colonDB}, lung infection~\cite{COVID}, breast lesions~\cite{BUSI}, and skin lesions~\cite{ISIC}. These tasks span multiple imaging modalities, including fundus imaging, MRI/CT, ultrasound, endoscopy, and dermoscopy. Due to substantial variations in lesion morphology, scale, texture, and imaging appearance, MLS remains a highly challenging problem that demands both robust generalization across modalities and fine-grained adaptability to specific lesion semantics.

\begin{figure}[t] 
    \centering
    \includegraphics[width=\columnwidth]{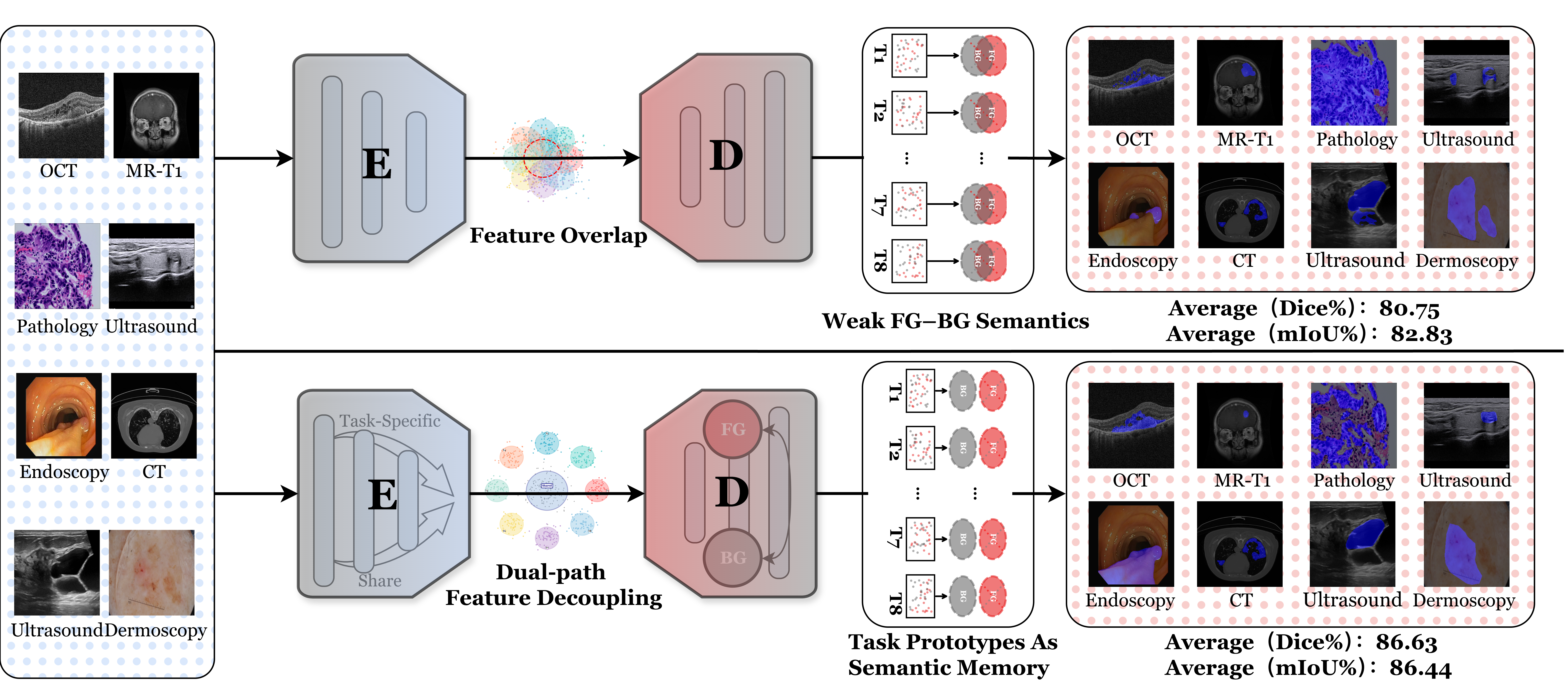}
    \caption{
    Comparison of framework design and average performance between previous unified models~\cite{SR-ICL,seggpt,spider,sam2unet} and our TP-Seg.  (1)  Single-path shared encoder vs. Dual-path routing encoder; (2) Task-agnostic decoder vs. Prototype-guided foreground-background decoder;  (3) Average Dice (\%): 80.75 vs. Average Dice (\%): 86.63;  (4)  Average mIoU (\%): 82.83 vs. Average mIoU (\%): 86.44. 
    % Comparison between conventional unified medical lesion segmentation frameworks and the proposed TP-Seg. Multi-task learning across diverse modalities suffers from feature entanglement and conflicting gradients. We propose TP-Seg, which decouples task-specific and shared representations through a dual-path expert module, and anchors task-specific semantics via learnable prototypes for improved foreground-background discrimination.
    % Due to the different modalities of images across the 8 tasks in unified MLS, previous methods tend to suffer from feature overlap and gradient interference during the encoding stage, and struggle to effectively distinguish foreground and background semantics during the decoding stage. TP-Seg addresses these issues by employing a dual-path expert module in the TCA to decouple shared and task-specific features, enabling cross-task knowledge sharing while maintaining task independence in the encoding stage; and by introducing task prototypes as persistent semantic memory in the decoding stage to effectively capture task-specific foreground and background semantics.
    }
    \label{fig:image1}
    \vspace{-5mm}
\end{figure}
Traditional deep learning methods~\cite{zhao1,decornet,egeunet,TRSRD-Net,cmunet,SOMANet} typically follow the single-task, single-model, single-set of parameters paradigm, where a dedicated model is trained for each lesion type. This approach, while effective for isolated tasks, leads to severe parameter redundancy, expensive training and deployment costs, and limited knowledge transfer among related tasks. To improve efficiency and scalability, recent studies~\cite{zhao7,universeg,SR-ICL,spider,seggpt} have explored unified medical lesion segmentation, aiming to use a single model and a single set of parameters to perform multiple segmentation tasks simultaneously. The primary advantage of this unified formulation lies not only in its training and deployment efficiency, avoiding repetitive task-specific model optimization, but also in its ability to aggregate diverse datasets during joint training. Such data unification acts as an implicit regularizer, alleviating the overfitting issues of small-scale lesion datasets and enhancing the model’s robustness and generalization capability.

As shown in Fig.~\hyperref[fig:image1]{1}, most existing unified segmentation approaches~\cite{SR-ICL,seggpt,universeg,spider,sam2-UNet} rely on a shared encoder to extract task-general features across heterogeneous datasets. This design overlooks the inherent differences in imaging modalities and lesion characteristics, leading to modality conflicts and representation entanglement. Inconsistent feature distributions across tasks cause overlapping gradients and interfere with effective optimization, ultimately degrading both overall and per-task segmentation performance. Moreover, conventional decoders are usually task-agnostic, meaning they treat all tasks uniformly without explicitly modeling task-specific semantics. The lack of task awareness makes it difficult to accurately capture distinct foreground–background relationships or adapt to the morphological variations of different lesion types. Consequently, these limitations result in semantic confusion, boundary inaccuracy, and inconsistent segmentation performance across tasks and modalities.
%
% Secondly, many traditional methods \cite{unet,transunet,sam2unet,SOMANet,zhao8,zhao9} use task-agnostic decoders, which lack sufficient task-awareness and fail to capture task-specific foreground/background semantics effectively. As a result, in multi-task scenarios, the decoder often cannot adjust according to the specific needs of each task during inference, leading to inaccurate boundary predictions, thus impacting segmentation accuracy and consistency across tasks.

To overcome these challenges, we introduce TP-Seg, a task-prototype framework designed for unified medical lesion segmentation. The design of TP-Seg is guided by two key observations: \noindent\textit{\textbf{\uppercase\expandafter{\romannumeral1})  }}
In multi-lesion learning, both shared and task-specific representations are crucial. Shared knowledge provides general visual priors, while task-specific encoding ensures modality and lesion sensitivity. 
\noindent\textit{\textbf{\uppercase\expandafter{\romannumeral2})  }}
Task-specific semantics should be explicitly represented and dynamically guided throughout the decoding process, rather than being implicitly learned from shared features. 
Based on these insights, we propose two complementary components. The task-conditioned adapter (TCA) is the first parameter-efficient fine-tuning (PEFT)-based module tailored for unified MLS. It introduces a dual-path expert design, where shared and task-specific experts jointly modulate feature extraction. This structure adaptively allocates representational capacity, mitigating feature interference while maintaining efficient knowledge sharing across tasks. The prototype-guided task decoder (PGTD) further enhances decoding by introducing learnable task prototypes as persistent semantic anchors. These prototypes evolve continuously during training, serving as a memory bank that captures task-specific foreground and background semantics, and guide the decoder via cross-attention to achieve consistent and precise predictions. Together, these designs enable TP-Seg to achieve both task adaptability and cross-task generalization within a unified framework.

% TP-Seg overcomes the shortcomings of preivous methods through two key components: 1) Task-Conditioned Adapter (TCA): Unlike previous shared adapter methods based on Parameter-Efficient Fine-Tuning (PEFT), we design the first PEFT approach specifically tailored for unified medical lesion segmentation.
% The proposed TCA introduces a shared and task-specific dual-path expert module, which effectively balances shared and task-specific encoding during feature extraction. This design enables the model to better handle modality discrepancies and feature conflicts in multi-task scenarios, while also fully leveraging shared common features across tasks to achieve optimal overall performance. 2) Prototype-Guided Task Decoder (PGTD): Traditional decoders often overlook task-specific semantic information, leading to imprecise decoding results. PGTD addresses this issue by introducing continuously updated task prototypes. Each task prototype serves as persistent semantic memory, accurately reflecting the underlying semantics of the task, and interacts with query features to guide task-specific predictions. Through this mechanism, PGTD enhances the model's cross-task generalization ability through the continuous evolution of task prototypes, further improving segmentation accuracy and consistency.

In summary, our main contributions are as follows:
\begin{itemize}
\item We introduce a simple yet effective task-prototype framework (TP-Seg), which unifies multi-lesion segmentation through task conditioning and prototype-guided semantic modeling.
% \item We design the Task-Conditioned Adapter, the first adapter specifically tailored for unified MLS tasks. It achieves a balance between shared and task-specific modeling through a dual-path expert architecture.
\item We design a parameter-efficient dual-path module that jointly models shared and task-specific representations.
% TCA dynamically allocates representational capacity during feature extraction, effectively mitigating modality conflicts and gradient interference while promoting cross-task feature sharing.
\item We propose a decoder guided by learnable task prototypes that serve as persistent semantic memory.
% \item We design the Prototype-Guided Task Decoder, which uses continuously updated task prototypes as persistent memory to remove reference dependence and enhance cross-task generalization.
\item Extensive experiments on 8 medical lesion benchmarks and multiple imaging modalities demonstrate that TP-Seg consistently achieves state-of-the-art results.
% \item We propose TP-Seg, a novel unified medical lesion segmentation framework that achieves SOTA performance on 8 MLS tasks. More importantly, TP-Seg establishes a new paradigm for generalized multi-lesion evaluation, allowing a single model trained once to effectively segment diverse lesion types across tasks and modalities. 
\end{itemize}

\section{Related Work}
\label{sec:formatting}
\begin{figure*}[t]
    \centering
    \includegraphics[width=0.9\textwidth]{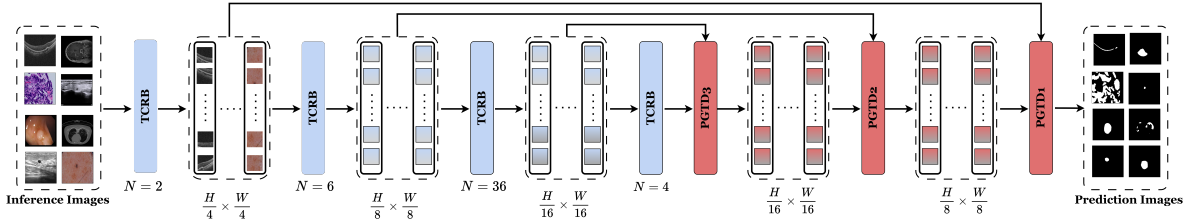}
    \caption{
    Overall architecture of the proposed TP-Seg framework for unified medical lesion segmentation. Each input image, together with its task embedding, is processed by the task-conditioned routing block (TCRB) for feature extraction, followed by the prototype-guided task decoder (PGTD) for task-aware decoding and final lesion prediction.
    }
    \label{fig:image2}
    \vspace{-3mm}
\end{figure*}

\begin{figure*}[htbp]
    \centering
    \includegraphics[width=0.9\textwidth]{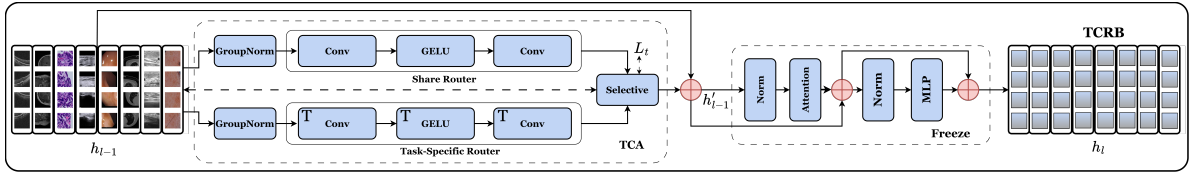}
    \caption{
    Illustration of the task-conditioned routing block (TCRB), which integrates a shared router and a task-specific router within each task-conditioned adapter (TCA) to achieve dynamic feature adaptation before the frozen encoder blocks.}
    % Schematic diagram of the Task-Conditioned Routing Block. A trainable Task-Conditioned Adapter (TCA) is inserted before each frozen encoder block, using a shared router and a task-specific router to balance shared and task-specific representations. The two branches produce feature increments $\Delta h_{s}$ and $\Delta h_{t}$, which are adaptively selected according to the task and block level, then added to the original input in a residual manner—enabling dynamic, task-aware modulation while preserving the frozen backbone.
    \label{fig:image3}
    \vspace{-3mm}
\end{figure*}

\subsection{Medical Lesion Segmentation}
%当前的医学病灶分割方法主要面向Specific病灶的专家模型，一般病灶的高泛化性通用模型以及高效的参数统一模型。
%介绍一些不同病灶的专家模型（主要总结出对各自病灶特点所提的专门设计）
%通用模型例如Unet, Unet++, M2SNet, TransUNet这些从病灶分割的通用问题出发，例如病灶多尺度大小，病灶隐藏属性的弱感知等问题~~
%统一模型。。。 在谈到visual in-context learning的时候，需要加上ICL的必要引用和定义介绍.... Different them, we maintain stable task-specific semantics through continuously updated task prototypes and enhance generalization via prototype-guided adaptive decoding.
Current medical lesion segmentation (MLS) methods can generally be categorized into three types: specialized expert models for specific lesions~\cite{LDNet,weakpolyp,infnet,decornet,malunet}, general models with high generalization capability~\cite{sam2-UNet,transunet,unet,swinunet}, and parameter-efficient unified models \cite{SR-ICL,spider,DSC-ICL,universeg,zhao10}.

\noindent \textbf{Specialized Models.}
%MSNet is designed for colon polyp segmentation, using multi-scale subtraction units to capture polyp-specific features and improve localization and edge clarity. AAU-Net and CMU-Net employ adaptive and multi-scale attention mechanisms to capture global contextual features and enhance boundary delineation in complex ultrasound images. Inf-Net and DECOR-Net employ attention and low-level feature enhancement to handle heterogeneous infections, fuzzy boundaries, and limited labels in COVID-19 CT segmentation.
%TRSRD-Net improves thyroid nodule segmentation using enhanced resolution and parallel atrous convolutions.
%需要写一些共性的方法，比如xx-based methods~\cite{msnet,msnet++,pranet...} utilize ... for addresing the ... of polyp. Some works~\cite{} adopt xx to percept the brain tumor... 以此类推。每个描述都一句话搞定，而不是对一个方法详细描述
%Inf-Net and DECOR-Net employ attention and low-level feature enhancement to handle heterogeneous infections, fuzzy boundaries, and limited labels in COVID-19 CT segmentation.
Some works~\cite{zhao1,weakpolyp,LDNet} based on scale-space theory employ multi-scale feature extraction strategies to alleviate boundary ambiguity and address limited annotation issues in polyp segmentation.
Approaches~\cite{aaunet,cmunet} grounded in adaptive and multi-scale attention mechanisms effectively capture global contextual information and enhance boundary delineation in complex breast ultrasound images.
For COVID-19 infection segmentation, attention-driven frameworks such as Inf-Net~\cite{infnet} and DECOR-Net~\cite{decornet} leverage low-level feature enhancement to handle heterogeneous infection patterns, fuzzy boundaries, and insufficient labeled data.
%TRSRD-Net~\cite{TRSRD-Net} further improves thyroid nodule segmentation by enhancing image resolution and employing parallel atrous convolutions to accurately delineate nodule boundaries.

\noindent \textbf{Generalized Models.}
%General models such as~\cite{unet,zhao4,unetplusplus} improve lesion perception and localization across scales by enhancing feature fusion and multi-scale learning. Some works~\cite{transunet,sam2-UNet} integrate Transformer architectures and pre-trained vision foundation models into U-shaped backbones to better exploit global context and facilitate knowledge transfer across modalities and tasks, thereby advancing generalized medical image segmentation frameworks.
%Architecture-refined U-shaped models such as~\cite{rolling,swinunet} redesign the token-mixing mechanism by introducing directional MLPs or windowed Transformer modules, enabling more efficient modeling of long-range dependencies under comparable computational budgets while preserving local context and improving performance on diverse medical image segmentation tasks.
UNet~\cite{unet} remains one of the most widely adopted architectures for medical lesion segmentation, featuring an encoder–decoder design. UNet++~\cite{unetplusplus} and M2SNet~\cite{zhao4} enhance lesion perception and localization by introducing dense multi-scale addition or subtraction operations across intra- and inter-layer connections. Recent works~\cite{transunet,sam2-UNet} integrate Transformer architectures or pre-trained vision models into U-shaped backbones to capture global context and facilitate knowledge transfer across tasks. Architecture-refined variants~\cite{rolling,swinunet} further improve token mixing through directional MLPs or windowed Transformer blocks, achieving stronger long-range dependency modeling and better segmentation performance with minimal computational overhead.

\noindent \textbf{Unified Models.}
In recent years, unified segmentation frameworks have shown great potential for MLS tasks. Inspired by In-Context Learning (ICL) from GPT-3~\cite{GPT3}, visual ICL has been explored in computer vision. MAE-VQGAN~\cite{WMDOVA} and Painter~\cite{point} achieve multi-task visual understanding via inpainting and masked modeling. SegGPT~\cite{seggpt} enables unified segmentation through random coloring, while UniverSeg~\cite{universeg}, Spider~\cite{spider}, and SR-ICL~\cite{SR-ICL} extend ICL to medical segmentation using reference-based or self-referring strategies. However, ICL-based methods degrade when reference-query matching is weak, leading to feature contamination. Unlike them, we maintain stable task-specific semantics via continuously updated task prototypes and improve generalization through prototype-guided adaptive decoding
%In contrast, TP-Seg maintains stable task-specific semantic representations by learning continuously updated task prototypes during training. Meanwhile, it achieves task-specific adaptation through prototype-guided attention and dynamic convolution generation, fundamentally enhancing the generalization and robustness of unified medical lesion segmentation.

\subsection{Parameter-Efficient Fine-Tuning}
Foundation models such as SAM \cite{sam} and SAM2 \cite{sam2} have demonstrated outstanding zero-shot segmentation capability through large-scale pretraining, providing strong general visual representations for downstream tasks \cite{zhao2,zhao4}. However, directly applying these models to unified medical lesion segmentation is not straightforward and typically requires PEFT to adapt them to downstream tasks. Typical methods such as Adapter \cite{adapter} and LoRA \cite{lora} reduce model size by sharing adapter parameters or low-rank subspaces across tasks. However, these shared-path designs face inherent challenges in unified medical lesion segmentation, as shared adapter cannot avoid conflicts caused by modality differences, leading to gradient interference between tasks. Therefore, we design TCA to address this pressing issue.
\section{Methodology}
% In this section, we first present the Method Overview in Sec. \ref{sec:3.1}. Then, we describe the two proposed components, Task-Conditioned Routing Block and Prototype-Guided Task Decoder, in Sec. \ref{sec:3.2} and Sec. \ref{sec:3.3}, respectively. Finally, the training and inference strategies are introduced in Sec. \ref{sec:3.4}.
% \subsection{Method Overview}
% \label{sec:3.1}

As shown in Fig.~\hyperref[fig:image2]{2}, TP-Seg is a unified framework for 8 medical lesion segmentation tasks. We employ SAM 2~\cite{sam2} as the encoder for feature extraction, where a task-conditioned adapter (TCA) is inserted before each of its 48 transformer blocks to form task-conditioned routing blocks (TCRB), enabling shared and task-specific adaptation. The decoder, composed of prototype-guided task decoder (PGTD) modules, integrates task prototypes, cross-attention, and dynamic convolution to guide task recognition and lesion segmentation efficiently across all tasks.

\subsection{Task-Conditioned Routing Block}
\label{sec:3.2}

As shown in Fig.~\hyperref[fig:image3]{3}, the input feature of the $l$-th encoder layer is denoted as $h_{l-1} \in \mathbb{R}^{B \times H \times W \times C}$ with task identifier ${\tau} \in \{1,\dots,T\}$. Each task-conditioned adapter (TCA) contains a shared router and multiple task-specific routers. The shared router extracts generic feature increments $\Delta h_s$ shared across all tasks, while the task-specific router computes task-dependent increments $\Delta h_t$. These two branches are fused by a selective gate to produce the adapted feature, which is then fed into the frozen SAM 2 transformer block for further processing. 
The feature increments are computed as:
\begin{equation}
	\Delta h_s = \text{Conv}_s\Big(\sigma\big(\text{Conv}_s(\text{GN}(h_{l-1}))\big)\Big),
\end{equation}

\begin{equation}
	\Delta h_t = \text{Conv}_t\Big(\sigma\big(\text{Conv}_t(\text{GN}(h_{l-1}))\big)\Big),
\end{equation}
where $\sigma(\cdot)$ denotes the GELU~\cite{GELU} activation function, $\text{GN}(\cdot)$ is the group normalization~\cite{GN},  $\text{Conv}_t(\cdot)$ and $\text{Conv}_s(\cdot)$ represent different convolution layers.

% Our core design principle is to enable all tasks to participate in shared adaptation as much as possible in the early stages of the network to learn generic visual representations, and then transition to independent adaptation in the later stages to capture task-specific discriminative features. Specifically, when a task selects shared adaptation, features propagate only through the shared router; when independent adaptation is selected, the system switches to the task-specific router. This staged routing strategy effectively satisfies the differentiated demands of different tasks for shared and independent adaptation, enabling the model to both preserve the generic visual priors captured by the shared router and focus on discriminative details through the task router.

To determine the optimal feature separation position $b_{\tau}^* \in \{2,...,47\}$ for each task, we introduce a learnable split gate vector  ${L}_{\tau} \in \mathbb{R}^{46}$ that dynamically controls the transition from shared to task-specific routing. For task ${\tau}$, the routing weight at block $b$ is computed using a cumulative sigmoid function:
\begin{equation}
	w_b^{\tau} = \sigma\left(\frac{\sum{i=2}^{b} {L}_{{\tau},i-2}}{tem}\right),
\end{equation}
where $tem$ is a temperature parameter linearly decaying from 1.0 to 0.3. This cumulative formulation enforces a monotonically increasing weight sequence, naturally producing a  ``shared-first, specific-later'' separation pattern. In the early high-temperature phase, the sigmoid remains smooth and allows gradients to explore potential separation points. As the temperature decreases, it becomes increasingly step-like and causes a sharp transition from $0 \rightarrow 1$ at a particular block position, which defines $b_{\tau}^*$. When $b < b_{\tau}^*$, shared features are used, and when $b \geq b_{\tau}^*$, task-specific adapters are activated. The entire process is end-to-end learnable. When a certain separation position improves validation performance, the corresponding gate parameters are updated through backpropagation, allowing each task to gradually converge to its optimal separation depth without manual tuning or discrete search.

\begin{figure}[t]
    \centering
    \includegraphics[width=\columnwidth]{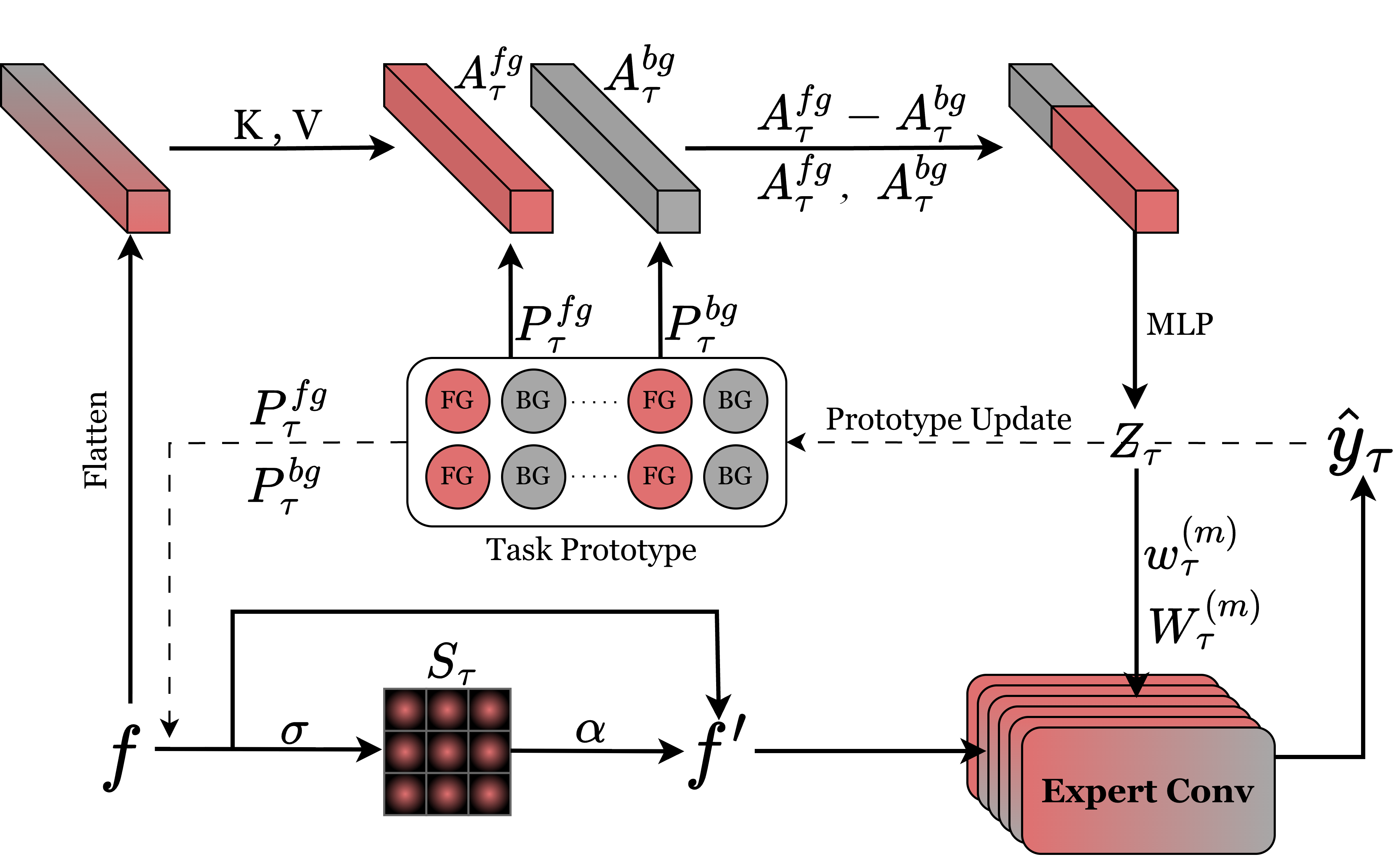}
    \caption{
    Illustration of the prototype-guided task decoder. 
%     Schematic diagram of the Prototype-Guided Task Decoder (PGTD). The fused feature \( f \) is flattened and processed through prototype cross-attention to generate the semantic descriptor \( Z_{\tau} \) for routing dynamic experts. Meanwhile, the prototype similarity map \( S_{\tau} \) modulates the feature \( f \) to produce \( f' \). The expert convolution applied to the modulated feature \( f' \) generates \( y_{\tau} \), which is then combined with prototype-guided reinforcement in the final prediction:
% $\hat{y}{\tau} = \text{Head}(y{\tau}) + \lambda_p \cdot \rho_{\tau} \cdot \tanh(S_{\tau}/\text{temp})$.
}
    \label{fig:image4}
    \vspace{-5mm}
\end{figure}

\subsection{Prototype-Guided Task Decoder}
\label{sec:3.3}
Although the TCRB effectively balances shared knowledge learning and task-specific adaptation through dual routing, the conventional decoder remains task-agnostic and fails to capture task-dependent foreground–background semantics. This limitation often results in imprecise boundaries and inconsistent predictions across tasks. Therefore, the proposed prototype-guided task decoder (PGTD) incorporates learnable task prototypes as persistent semantic anchors. These prototypes interact with spatial features via cross-attention to capture task-aware semantics, while dynamic convolutional experts and prototype-guided similarity maps further enable adaptive feature modulation, achieving unified yet precise multi-task lesion segmentation.

\noindent \textbf{Feature Fusion.}
Each PGTD module receives a pair of hierarchical encoder features: the low-level feature 
$f_{\text{low}} \in \mathbb{R}^{B \times C_{\text{low}} \times H \times W}$ 
and the high-level feature 
$f_{\text{high}} \in \mathbb{R}^{B \times C_{\text{high}} \times H' \times W'}$.  
To align their semantic and channel spaces, both are first normalized and projected through $1\times1$ convolutions, then fused as:
\begin{equation}
f = \text{C}\big(
\sigma(\text{GN}(\text{Conv}_{1\times1}(f_{\text{low}}))),
\sigma(\text{GN}(\text{Conv}_{1\times1}(f_{\text{high}})))
\big),
\end{equation}
where $\text{C}(\cdot)$ is the channel-wise concatenation operation. 
% Each PGTD receives two hierarchical feature maps from the encoder the low-level feature $f_{\text{low}} \in \mathbb{R}^{B \times C_{\text{low}} \times H \times W}$ and the high-level feature $f_{\text{high}} \in \mathbb{R}^{B \times C_{\text{high}} \times H' \times W'}$. To align their semantic space and channel dimensionality, both inputs are first processed through $1\times1$ convolutions followed by normalization and activation, then fused as $f = \text{fuse}( \sigma\big(\text{GN}(\text{Conv}_{1\times1}(f_{\text{low}}))\big)
% , \sigma\big(\text{GN}(\text{Conv}_{1\times1}(f_{\text{high}}))\big))$

\noindent \textbf{Prototype Interaction.}
As shown in Fig.~\hyperref[fig:image4]{4}, each task $\tau \in \mathcal{T}$ maintains a pair of foreground and background prototypes $\{P_{\tau}^{\text{fg}}, P_{\tau}^{\text{bg}}\}$, initialized as:
\begin{equation}
[P_{\tau}^{\text{fg}}, P_{\tau}^{\text{bg}}] = \text{MLP}(\tau),
\end{equation}
where MLP$(\cdot)$ denotes multilayer perceptron \cite{MLP}.
To establish task-aware semantic correspondence, the prototype queries attend spatial features via cross-attention:
\begin{equation}
A_{\tau}^{\text{fg}} = \text{Attn}(P_{\tau}^{\text{fg}}, K, V), \quad
A_{\tau}^{\text{bg}} = \text{Attn}(P_{\tau}^{\text{bg}}, K, V),
\end{equation}
where $K, V = \text{Flatten}(f)$, and Attn$(\cdot)$ is the cross-attention mechanism \cite{attention}.  
The responses are then aggregated into a discriminative semantic descriptor:
\begin{equation}
Z_{\tau} = \text{MLP}\big([A_{\tau}^{\text{fg}}, A_{\tau}^{\text{bg}}, A_{\tau}^{\text{fg}} - A_{\tau}^{\text{bg}}]\big),
\end{equation}
which captures both shared and contrasting semantic cues between the two regions.
% As illustrated in Fig.~\hyperref[fig:image4]{4}, each task $\tau \in \mathcal{T}$ maintains a pair of foreground and background prototypes $\{P_{\tau}^{\text{fg}}, P_{\tau}^{\text{bg}}\}$, initialized via $[P_{\tau}^{\text{fg}}, P_{\tau}^{\text{bg}}] = \text{MLP}(\tau)$. To introduce task-aware semantic interaction, prototype queries attend spatial features through cross-attention: $A_{\tau}^{\text{fg}}= \text{Attn}(P_{\tau}^{\text{fg}}, K, V)$ and $A_{\tau}^{\text{bg}}= \text{Attn}(P_{\tau}^{\text{bg}}, K, V)$, where $K, V = \text{Flatten}(f)$. These responses are fused to form a task-conditioned semantic descriptor:
% \begin{equation}
% Z_{\tau} = \text{MLP}\big([A_{\tau}^{\text{fg}}, A_{\tau}^{\text{bg}}, A_{\tau}^{\text{fg}}-A_{\tau}^{\text{bg}}]\big),
% \end{equation}
% which captures discriminative foreground and background cues for each task.

\noindent \textbf{Dynamic Experts and Feature Modulation.}
Based on $Z_{\tau}$, PGTD generates $M$ dynamic convolutional experts $\{W_{\tau}^{(m)}\}_{m=1}^{M}$ 
and routing weights $\{w_{\tau}^{(m)}\}_{m=1}^{M}$.  
Simultaneously, a prototype-guided similarity map is computed as:
\begin{equation}
S_{\tau} = \sigma(\langle f, P_{\tau}^{\text{fg}} \rangle - \langle f, P_{\tau}^{\text{bg}} \rangle),
\end{equation}
where $\langle \cdot, \cdot \rangle$ denotes cosine similarity.  
The modulated feature is then obtained by:
\begin{equation}
f' = f \cdot (1 + \alpha S_{\tau}),
\end{equation}
which is adaptively processed through the expert ensemble to yield the task-specific response:
\begin{equation}
y_{\tau} = \sum_{m=1}^{M} w_{\tau}^{(m)} \cdot (W_{\tau}^{(m)} * f').
\end{equation}

% To enhance task specificity, PGTD generates dynamic convolutional experts $\{W_{\tau}^{(m)}\}_{m=1}^{M}$ and routing weights $\{w_{\tau}^{(m)}\}_{m=1}^{M}$ from $Z_{\tau}$. In parallel, the prototype similarity map is computed to modulate features:
% \begin{equation}
% S_{\tau} = \sigma(\langle f, P_{\tau}^{\text{fg}} \rangle - \langle f, P_{\tau}^{\text{bg}} \rangle) \in [0,1]^{H \times W},
% \end{equation}
% where $\langle \cdot, \cdot \rangle$ denotes cosine similarity and $\sigma$ is sigmoid activation.
%  The modulated features $f' = f \cdot (1 + \alpha S_{\tau})$ are processed by experts to produce task-specific outputs: $y_{\tau} = \sum_{m=1}^{M} w_{\tau}^{(m)} \cdot (W_{\tau}^{(m)} * f')$.

\noindent \textbf{Prototype-Guided Reinforcement.}
Finally, the overall prediction is obtained by combining the expert output with a prototype-guided reinforcement term:
\begin{equation}
\hat{y}_{\tau} = \text{Head}(y_{\tau}) + \lambda_{p} \cdot \rho_{\tau} \cdot \tanh(S_{\tau} / \text{temp}),
\end{equation}
where $\lambda_{p}=5.0$ controls the reinforcement strength, $\rho_{\tau}$ is a learnable amplifier, and $\text{temp}$ adjusts the sharpness of the guidance. 
Through this joint mechanism, PGTD establishes strong semantic alignment between prototypes and spatial features, effectively capturing discriminative cues and ensuring precise, task-consistent lesion segmentation within a unified framework.

% The final prediction combines expert output with prototype-guided reinforcement:
% \begin{equation}
% \hat{y}_{\tau} = \text{Head}(y_{\tau}) + \lambda_{p} \cdot \rho_{\tau} \cdot \tanh(S_{\tau}/\text{temp}).
% \end{equation}
% where $\lambda_{p}=5.0$ is a fixed weight, $\rho_{\tau}$ is a learnable amplifier, and $\text{temp}$ controls guidance sharpness.

\noindent \textbf{Prototype Update Strategy.}
During training, each task $\tau$ maintains two prototypes, representing the foreground and background feature centers. These prototypes are updated using an exponential moving average (EMA)~\cite{EMA} to continuously capture the evolving feature distributions. At each iteration, the prototypes are updated as follows:
\begin{equation}
P_{\tau}^{*} \leftarrow m P_{\tau}^{*} + (1 - m) \hat{f}_{\tau}^{*}, \quad * \in \{\text{fg}, \text{bg}\},
\end{equation}
where $m = 0.9$ is the momentum coefficient and $\hat{f}_{\tau}^{*}$ denotes the mean feature extracted from the corresponding ground-truth region.
Given a feature map $f$ and its binary mask $M^{*}$, the mean feature $\hat{f}_{\tau}^{*}$ is computed as:
\begin{equation}
\hat{f}_{\tau}^{*} = \text{Normalize}\left(\frac{\sum_{i,j} M_{i,j}^{*} \cdot f_{i,j}}{\sum_{i,j} M_{i,j}^{*} + \epsilon}\right),
\end{equation}
%Todo i,j
where the normalization operation ensures that all prototypes lie on the unit hypersphere, enabling stable cosine similarity computation.
This momentum-based update mechanism effectively balances historical and current feature information. It prevents abrupt prototype shifts caused by noisy samples while allowing gradual adaptation to evolving feature distributions. All prototypes are initialized with task embeddings and are refined through these gradient-free EMA updates whenever the corresponding task is encountered during training.

% During training, task-specific prototypes are maintained and updated via EMA~\cite{EMA} to capture the evolving feature distributions. For each task $\tau$, we maintain both foreground and background prototypes that are updated using L2-normalized mean features extracted from ground-truth mask regions: $P_{\tau}^{*} \leftarrow m P_{\tau}^{*} + (1-m) \hat{f}_{\tau}^{*}, \quad * \in \{\text{fg}, \text{bg}\}$, where $m=0.9$ is the momentum coefficient and $\hat{f}_{\tau}^{*}$ represents the spatially-averaged feature within the corresponding mask region. Specifically, given the feature map $f$ and binary mask $M^{*}$, the mean feature is computed as: $\hat{f}_{\tau}^{*} = \text{Normalize}\left(\frac{\sum_{i,j} M_{i,j}^{*} \cdot f_{i,j}}{\sum_{i,j} M_{i,j}^{*} + \epsilon}\right)$, where the normalization ensures prototypes remain on the unit hypersphere for stable cosine similarity computation. The momentum-based update balances historical information with current observations, preventing drastic changes from individual noisy samples while allowing gradual adaptation to feature distribution shifts. Prototypes are initialized using task embeddings and subsequently refined through gradient-free EMA updates at each training iteration when the corresponding task is encountered.

\subsection{Training and Inference}
\label{sec:3.4}
\paragraph{Training.}
We follow a progressive joint optimization strategy to balance multiple lesion segmentation tasks within a single model. Since datasets vary significantly in scale, with wet AMD containing about five times more samples than breast lesion, we use a WeightedRandomSampler~\cite{weightedrandomsampling} to balance task occurrence at the batch level. For each task $\tau$ (${\tau}=0,\dots,n$), we compute its sample count $N_{\tau}$, derive inverse-frequency weights $w_{\tau} = 1/N_{\tau}$, and assign sampling probabilities as $P({\tau}_i) = w_{{\tau}_i} / \sum_j w_{{\tau}_j}$. This sampling strategy ensures that each task maintains a relatively stable representation ratio in training, which alleviates gradient bias toward large datasets and promotes more balanced optimization across tasks.

\noindent \textbf{Inference.}
We adopt a task-specific inference protocol in which the model processes one dataset at a time according to its task identifier (${\tau}=0,\dots,n$), and retrieves the corresponding task prototypes from trained weights. This setting emulates single-task inference while preserving the unified model structure, allowing us to clearly assess the performance of each individual task.

\section{Experiment}
\label{experiment}
\subsection{Experimental Setup}
\paragraph{Datasets and Evaluation Metrics.}
The dataset information for the 8 selected medical lesion segmentation tasks is shown in Tab.~\hyperref[tab:1]{1}. AMD-SD \cite{wetamd} annotates five pathological manifestations of wet AMD, which we unify into a single category for wet AMD lesion segmentation. BTD \cite{BTD1, BTD2} is divided into four folds for cross-validation; in our experiments, the first three folds are used for training and the fourth for validation. For EBHI-Seg \cite{ebhi}, we split the ADC segmentation subset into training and validation sets with a ratio of 4:1. We adopt the widely used Dice score and mean Intersection-over-Union (mIoU) for quantitative evaluation, where higher values indicate better performance.

\begin{table}[t]
\centering
\resizebox{\columnwidth}{!}{%
\begin{tabular}{r|cccc}
\specialrule{1.5pt}{0pt}{0pt}
Segmentation Task & Dataset & Modality & \#Train & \#Val \\
\midrule
Wet AMD (WA)& AMD-SD \cite{wetamd} & OCT & 2,346 & 703 \\
Brain Tumor (BT)& BTD \cite{BTD1,BTD2}& MR-T1 & 2,298 & 766 \\
Adenocarcinoma (ADC)& EBHI-Seg \cite{ebhi} & Pathology image & 636 & 159 \\
Thyroid Nodule (TN)& TNUI 2021 \cite{TNUI} & Ultrasound & 966 & 276 \\
Colon Polyp (CP)& Five datasets \cite{polyp1,polyp2,polyp3,polyp4,polyp5} & Endoscopy image & 1,450 & 798 \\
Lung Infection (LI)& COVID-19 data \cite{COVID} & CT & 894 & 385 \\
Breast Lesion (BL)& BUSI \cite{BUSI} & Ultrasound & 486 & 161 \\
Skin Lesion (SL)& ISIC 2018 \cite{ISIC} & Dermoscopy image & 1,886 & 808 \\
\specialrule{1.5pt}{0pt}{0pt}
\end{tabular}%
\label{tab:1}
}
\caption{Data partition of the 8 selected medical lesion segmentation tasks, commonly used by state-of-the-art specialized and unified methods.}
\vspace{-4mm}
\end{table}
\begin{table*}[t]
    \centering 
    \setlength{\tabcolsep}{0.4mm}
    \resizebox{0.9\linewidth}{!}{\begin{tabular}{r|c|cc|cc|cc|cc|cc|cc|cc|cc|cc}
\specialrule{1.5pt}{0pt}{0pt}
\toprule
\multirow{2}{*}{Methods}
& \multirow{2}{*}{Publication}
& \multicolumn{2}{c|}{Wet AMD}  
& \multicolumn{2}{c|}{Brain Tumor} 
& \multicolumn{2}{c|}{Adenocarcinoma}  
& \multicolumn{2}{c|}{Thyroid Nodule}
& \multicolumn{2}{c|}{Colon Polyp}
& \multicolumn{2}{c|}{Lung Infection}
& \multicolumn{2}{c|}{Breast Lesion} 
& \multicolumn{2}{c|}{Skin Lesion}
& \multicolumn{2}{c}{Average} \\
&
& Dice $\uparrow$ & mIoU $\uparrow$
& Dice $\uparrow$ & mIoU $\uparrow$
& Dice $\uparrow$ & mIoU $\uparrow$
& Dice $\uparrow$ & mIoU $\uparrow$
& Dice $\uparrow$ & mIoU $\uparrow$
& Dice $\uparrow$ & mIoU $\uparrow$
& Dice $\uparrow$ & mIoU $\uparrow$
& Dice $\uparrow$ & mIoU $\uparrow$
& Dice $\uparrow$ & mIoU $\uparrow$
\\
\hline 
\rowcolor[rgb]{0.500,0.965,0.973}\multicolumn{20}{c}{\textbf{Specialized and General Models (One model for one task)}} \\
\hline 
 TRSRD-Net \cite{TRSRD-Net} &ICISN'24 & - & -& - & -& -& -& 85.40& 84.65 & -& -& -& -& - & -& - & -& - & - \\
 LDNet \cite{LDNet}& MICCAI'23 & -& - & - & -& - & -& - & -& 64.25& 74.41& -& -& -& -& - & -& -& - \\
WeakPolyp \cite{weakpolyp} & MICCAI'23 & - & - & - & - & - & - & - & - & 74.90& 80.66 & - & -& - & - & - & - & - & - \\
Inf-Net \cite{infnet} & TMI'23 & - & - & - & - & -& - & - & - & - & -& 43.24 &52.85 & - & -&- & - & - \\
DECOR-Net \cite{decornet}& ISBI'23& - & - & - & - & - & - & - & - & - & -& 40.25 &69.49 & - & - & -&- & - & - \\
AAU-Net \cite{aaunet}& TMI'22 & - & - & - & - & -& -& - & - & - & - & -& -& 47.45 & 65.15 & - & - & -& - \\
CMU-Net \cite{cmunet} & ISBI'23 & - & - & - & - & -& -  & - & - & - & - & -& -  & 54.52& 83.02 & -& - & - \\
MALUNet \cite{malunet} & BIBM'22 & - & - & - & - & -& - & - & - & - & - & - & -& - & - & 86.32 & 85.37 & -& -  \\
EGE-UNet \cite{egeunet} & MICCAI'23 & - & - & - & - & -& - & - & - & - & - & - & -& - & - & 85.88 & 84.98 & -& -   \\
TransUNet \cite{transunet}& MIA'24 & 76.72& 81.29 & 57.92 & 72.46 & 91.85 & 74.23& 78.24 & 82.90& 66.41& 75.52 & 43.89& 65.33 & 75.67 & 80.62& 86.40 & 82.53& 72.14& 76.86 \\
Spider \cite{spider} & ICML'24  & 79.32& 83.22 & 72.94& 80.17 & \color{mygreen}\textbf{93.74} & 80.31& \color{mygreen}\textbf{86.39}& 85.24 & 81.22& \color{mygreen}\textbf{85.83} & 75.21 & 81.89& 82.31& 84.58 & 89.39 & 86.16 & 82.56 & 83.42 \\
SAM2-UNet \cite{sam2-UNet} & ICCVW'25  & \color{mygreen}\textbf{83.13}& \color{mygreen}\textbf{84.21} & \color{mygreen}\textbf{73.71}& 78.92 & 93.51& \color{mygreen}\textbf{84.16} & 85.52& 85.37 & \color{mygreen}\textbf{83.28}& 83.22 & \color{mygreen}\textbf{82.34} & 83.57 & \color{mygreen}\textbf{83.39}& 83.36 & 89.12 & \color{mygreen}\textbf{88.24}& \color{mygreen}\textbf{84.25} & 84.13\\
SR-ICL \cite{SR-ICL} & CVPR'25 & 80.70& 83.88 & 70.90& 79.26 & 93.46& 80.28 & 86.52& \color{reda}\textbf{88.03} & 81.48& 85.70 & 79.89& 84.65 & 83.06& \color{reda}\textbf{85.56} & \color{mygreen}\textbf{89.73}& 86.63 & 83.23& \color{mygreen}\textbf{84.24}\\
\rowcolor[rgb]{0.500,0.965,0.973}TP-Seg-General & -& \color{reda}\textbf{86.67} & \color{reda}\textbf{88.62}& \color{reda}\textbf{75.65} &\color{reda}\textbf{81.03}& \color{reda}\textbf{94.80} & \color{reda}\textbf{85.55} & \color{reda}\textbf{86.58}& \color{mygreen}\textbf{86.37} & \color{reda}\textbf{86.66} & \color{reda}\textbf{87.28}& \color{reda}\textbf{85.91} & \color{reda}\textbf{87.03}& \color{reda}\textbf{85.06}& \color{mygreen}\textbf{85.11} & \color{reda}\textbf{90.00}& \color{reda}\textbf{88.47} & \color{reda}\textbf{86.41}& \color{reda}\textbf{86.18}\\
\hline 
\rowcolor[rgb]{0.500,0.965,0.973}\multicolumn{20}{c}{\textbf{Unified Models (One model for all tasks)}} \\
\hline 
SegGPT \cite{seggpt}& ICCV'23 & 71.66& 78.64 & 44.17 & 67.30 & 94.24 & 80.14 & 79.69 & 85.18 & 78.29 & 84.66 & 50.10 & 65.91 & 78.87& 83.87 & 87.90 & 85.42& 73.11 & 78.89 \\
% UniverSeg \cite{universeg}& ICCV'23 &  45.90 & 63.66  &32.83  &61.29  &80.85  &45.66  &60.95  &73.63  &29.31  &53.46  &39.98  &65.21  &65.51  &74.16  &77.16  &76.01 & 54.06 & 78.89 \\

% TransUNet \cite{transunet}& MIA'24 & 75.46& 80.43 & 56.59 & 71.77& 90.34& 69.16  & 78.57& 83.21 & 55.81& 66.21 & 36.68 & 61.62& 74.66 & 79.53& 86.14 & 80.67 & 68.90& 74.07 \\
Spider \cite{spider} & ICML'24   & 78.51 & 82.66 & 71.59 & 79.86 & 93.65 & 80.30 & 86.52 & 88.40 & 80.66& 85.30 & 73.85 & 82.07 & 81.16 & 84.80 & 88.79& 87.18  & 81.84 & 83.82 \\
SAM2-UNet \cite{sam2-UNet} & ICCVW'25 & \color{mygreen}\textbf{82.49}& \color{mygreen}\textbf{84.33}& 73.57 & 78.64 & 92.35 & \color{mygreen}\textbf{82.31}& 84.77& 84.69 & 82.47& 82.33& \color{mygreen}\textbf{82.65} & 83.06& 82.79 & 83.16 & 88.31 & 87.07 & 83.67& 83.19 \\
SR-ICL \cite{SR-ICL} & CVPR'25 & 80.54 & 83.84 & \color{mygreen}\textbf{74.29} & \color{mygreen}\textbf{80.50} & \color{mygreen}\textbf{94.96} & 81.98 & \color{reda}\textbf{87.91}& \color{reda}\textbf{89.18} & \color{mygreen}\textbf{83.26} & \color{mygreen}\textbf{86.53} & 82.36&\color{mygreen}\textbf{87.21} & \color{mygreen}\textbf{84.92}& \color{reda}\textbf{87.08} & \color{reda}\textbf{90.85} & \color{mygreen}\textbf{87.29} & \color{mygreen}\textbf{84.88}& \color{mygreen}\textbf{85.45} \\

\rowcolor[rgb]{0.500,0.965,0.973}TP-Seg-Unified & - & \color{reda}\textbf{86.53}& \color{reda}\textbf{88.50} & \color{reda}\textbf{75.23}& \color{reda}\textbf{80.56} & \color{reda}\textbf{95.05}& \color{reda}\textbf{85.79} & \color{mygreen}\textbf{87.66}& \color{mygreen}\textbf{87.53} &\color{reda}\textbf{86.48}& \color{reda}\textbf{86.59} &\color{reda}\textbf{86.47} &\color{reda}\textbf{88.17} & \color{reda}\textbf{85.49}& \color{mygreen}\textbf{85.66} & \color{mygreen}\textbf{90.15}&\color{reda}\textbf{88.69} &\color{reda}\textbf{86.63}&\color{reda}\textbf{86.44}\\
\specialrule{1.5pt}{0pt}{0pt}
\end{tabular}
}
\label{tab:2}
\caption{Quantitative comparison with state-of-the-art methods in terms of Dice and mIoU (\%). $\uparrow$ indicates that larger scores are better. For specialized models, each task is trained independently, while unified models are trained jointly across the 8 MLS tasks. ``Average'' denotes the mean performance across all 8 tasks.  The top two results are highlighted in {\color{mygreen}\textbf{green}} and {\color{reda}\textbf{red}}.
}
\end{table*}

\noindent \textbf{Implementation details.}
TP-Seg is implemented in PyTorch~\cite{pytorch}  and trained on an NVIDIA RTX 5090 GPU. We use the Adam optimizer~\cite{adam} with an initial learning rate of 3e-5 and train the model for 30 epochs until convergence. For backbone initialization, we adopt the Hiera-L~\cite{sam2} variant of SAM 2 pretrained weights to accelerate convergence and enhance overall performance.

\subsection{Quantitative Evaluation}
Tab.~\hyperref[tab:2]{2} reports the quantitative comparison across 8 medical lesion segmentation tasks in terms of Dice and mIoU scores. We present two variants of our framework: TP-Seg-General, where models are trained independently for each task, and TP-Seg-Unified, where a single model is jointly trained across all tasks. 
Compared to specialized expert models and the general-purpose TransUNet, TP-Seg-General achieves the highest or second-highest performance on nearly all tasks, with an average Dice of 86.42\%.
When jointly trained across 8 tasks, TP-Seg-Unified further enhances generalization, achieving an average Dice of 86.63\%, outperforming all unified baselines. In particular, it consistently surpasses SegGPT~\cite{seggpt}, SAM2-UNet~\cite{sam2-UNet}, and SR-ICL~\cite{SR-ICL} across multiple lesion categories, indicating its superior capability in cross-task feature sharing and robust task-conditioned adaptation.
Moreover, we retrain Spider, SAM2-UNet~\cite{sam2unet}, and SR-ICL~\cite{SR-ICL} in a task-wise manner under the same settings, and their performances remain inferior to our TP-Seg-General. This verifies that our framework’s advantage stems not merely from training strategy, but from its structural generality and explicit task-conditioning mechanism, enabling it to maintain strong performance in both specialized and unified scenarios.

\begin{figure*}[htbp]
    \centering
    \includegraphics[width=0.9\linewidth]{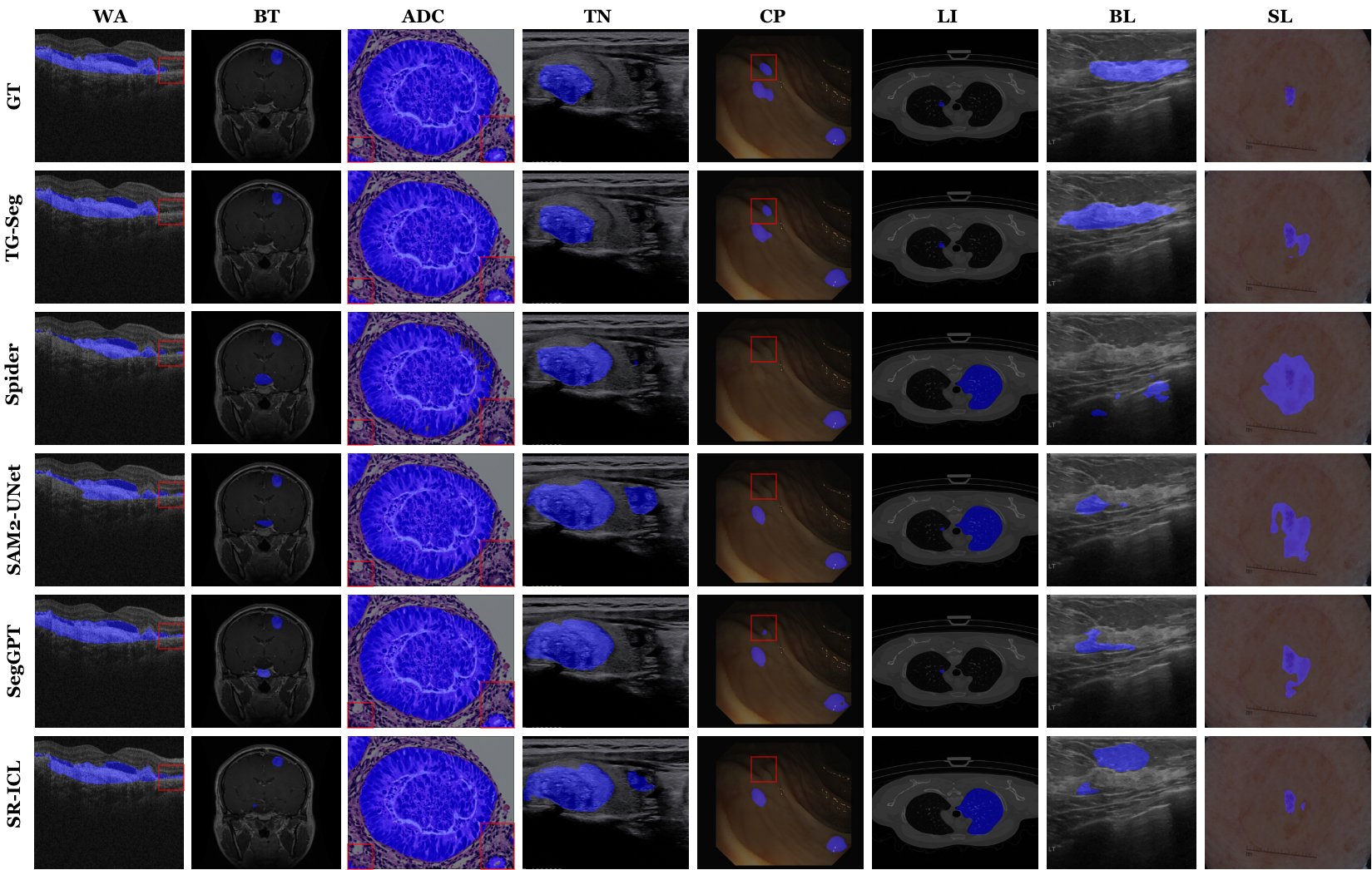}
 \caption{Visual comparison of TP-Seg with other unified models, including Spider~\cite{spider}, SAM2-UNet~\cite{sam2-UNet} ,SegGPT~\cite{seggpt} and SR-ICL~\cite{SR-ICL}, across the 8 medical lesion segmentation tasks.}

    \label{fig:image5}
    \vspace{-3mm}
\end{figure*}

\begin{table*}[t]
    \centering 
    \setlength{\tabcolsep}{0.4mm}
    \resizebox{0.9\linewidth}{!}{\begin{tabular}{cc|cc|cc|cc|cc|cc|cc|cc|cc|cc}
\specialrule{1.5pt}{0pt}{0pt}
\toprule
\multirow{2}{*}{ID}
& \multirow{2}{*}{Settings}
& \multicolumn{2}{c|}{Wet AMD}  
& \multicolumn{2}{c|}{Brain Tumor} 
& \multicolumn{2}{c|}{Adenocarcinoma}  
& \multicolumn{2}{c|}{Thyroid Nodule}
& \multicolumn{2}{c|}{Colon Polyp}
& \multicolumn{2}{c|}{Lung Infection}
& \multicolumn{2}{c|}{Breast Lesion} 
& \multicolumn{2}{c|}{Skin Lesion}
& \multicolumn{2}{c}{Average} \\
&
& Dice $\uparrow$ & mIoU $\uparrow$
& Dice $\uparrow$ & mIoU $\uparrow$
& Dice $\uparrow$ & mIoU $\uparrow$
& Dice $\uparrow$ & mIoU $\uparrow$
& Dice $\uparrow$ & mIoU $\uparrow$
& Dice $\uparrow$ & mIoU $\uparrow$
& Dice $\uparrow$ & mIoU $\uparrow$
& Dice $\uparrow$ & mIoU $\uparrow$
& Dice $\uparrow$ & mIoU $\uparrow$
\\
\hline 
\rowcolor[rgb]{0.500,0.965,0.973}\multicolumn{20}{c}{(A) Each Component } \\
\hline 
A1 & Baseline  & 81.29 &82.42  & 66.39& 73.96 & 92.47 & 82.28 & 79.83 &78.67 & 63.71& 64.58 & 80.79 &81.74  & 78.42 &78.19 & 86.64 & 82.39 & 78.69 & 78.02  \\
A2 & w/o TCA  & 83.36& 84.55 & 68.78& 76.21 & 93.76&83.22  & 80.65& 79.74 & 65.85 &66.17 & 81.37 &82.44  & 79.65& 79.43 & 87.21& 82.87  & 80.08&79.32   \\
A3 & w/o PGTD  & 85.37 & 85.95 & 74.22& 79.17 & 94.63& 84.21 & 86.83& 85.63 &85.33 & 85.51 & 84.72&86.39  & 84.17 &84.23  & 89.11& 87.77 & 85.54&84.85  \\
\rowcolor[rgb]{0.500,0.965,0.973}A4 & Full (TP-Seg)  & 86.53 & 88.50 & 75.23 &80.56  & {95.05} &{85.79} & {87.66} &{87.53} &86.48 & 86.59& {86.47} & {88.17} & {85.49} &{85.66} & {90.15}&{88.69} & {86.63}& {86.44}\\

A5 & Separate Training  & {86.67} &{88.62} & {75.65} &{81.03} & 94.80& 85.55 & 86.58&86.37  & {86.66}& {87.28} & 85.91 &87.03 & 85.06&85.11  & 90.00&88.47  & 86.41&86.18 \\
\hline 
\rowcolor[rgb]{0.500,0.965,0.973}\multicolumn{20}{c}{(B) Task-Conditioned Adapter} \\
\hline 
B1 & LoRA & 85.44 &86.01 & 74.02 & 79.03 & 94.74 & 84.31 & 86.57 &86.35& 84.74 &84.96 & 84.49 &86.27& 84.63&84.74 & 89.21&87.85 & 85.48&84.94\\
B2 & Adapter & 85.31& 85.99 & 74.25&79.30 & 94.69 &84.30 & 86.53 &86.31 & 84.82&85.01 & 84.45 &86.03 & 84.61 &84.65 & 89.32 &87.93 & 85.49&84.83\\
B3& w/o Task-Specific Route & 85.56&86.52 & 72.88 &79.01& 94.47 &84.19 & 85.59&85.74 & 84.06&84.33 & 84.94&86.69 & 85.57 & 85.70& 89.35&87.96 & 85.30&85.01\\
B4& w/o Share Route & 86.35&88.24 & 75.31 &80.63& 95.02&85.70 & 87.72 &87.61&86.12&86.20& 85.59& 87.13& 84.97 &85.02& 90.10&88.65& 86.40&86.14\\
\hline 
\rowcolor[rgb]{0.500,0.965,0.973}\multicolumn{20}{c}{(C) Hyperparameters} \\
\hline
C1 & $\lambda_{p}=2.5$,$ \alpha=0.5$,Experts M=6 & 86.33&88.20 & 74.97&80.24 & 94.98&85.65 & 87.36&87.20 & 85.99&86.03 & 85.61& 87.16& 85.12 &85.21& 89.89 &88.46& 86.28 &86.01 \\
C2 & $\lambda_{p}=10$,$ \alpha=0.5$,Experts M=6 & 86.20 &88.13& 75.02&80.31  & 94.84&85.47 & 87.05 &86.89& 86.54&86.60 & 86.37&88.02 & 85.15 &85.26& 89.84&88.41 & 86.37 & 86.13\\
C3& $\lambda_{p}=5.0$,$ \alpha=0.6$,Experts M=8  & 86.26& 88.17& 75.27 &80.59& 95.00&85.71 & 87.63 &87.50&86.58&86.64& 85.59&87.11 & 85.31 &85.44& 90.02&88.58& 86.45&86.21 \\
C4& $\lambda_{p}=5.0$,$ \alpha=0.5$,Experts M=4   & 86.31& 88.23& 74.17 &79.26& 94.53& 85.25& 87.63&87.51 & {86.89} &{86.98}& 85.48&87.09  & 85.06 &85.15& 89.67 &88.24& 86.22&85.96  \\
C5& $\lambda_{p}=5.0$,$ \alpha=0.5$,Experts M=8  & 85.92 &87.87 & 74.76& 79.98& 94.79&85.40 & 87.29&87.13 & 86.44& 86.51& 85.39&87.00  & {85.52}&{ 85.71}& 89.88& 88.43& 86.24& 86.00 \\
\rowcolor[rgb]{0.500,0.965,0.973}C6 &$\lambda_{p}=5.0$,$ \alpha=0.5$,Experts M=6 & {86.53}&{88.50} & {75.23} &{80.56} & {95.05} &{85.79}& {87.66} &{87.53} &86.48&86.59 & {86.47}&{88.17} & 85.49&85.66 & {90.15}&{88.69}& {86.63}&{86.44}\\
\specialrule{1.5pt}{0pt}{0pt}
\end{tabular}
}
\label{tab:3}
\caption{Ablation studies on the 8 MLS tasks.}
\end{table*}

\begin{figure}[htbp]
    \centering
    \includegraphics[width=0.95\columnwidth]{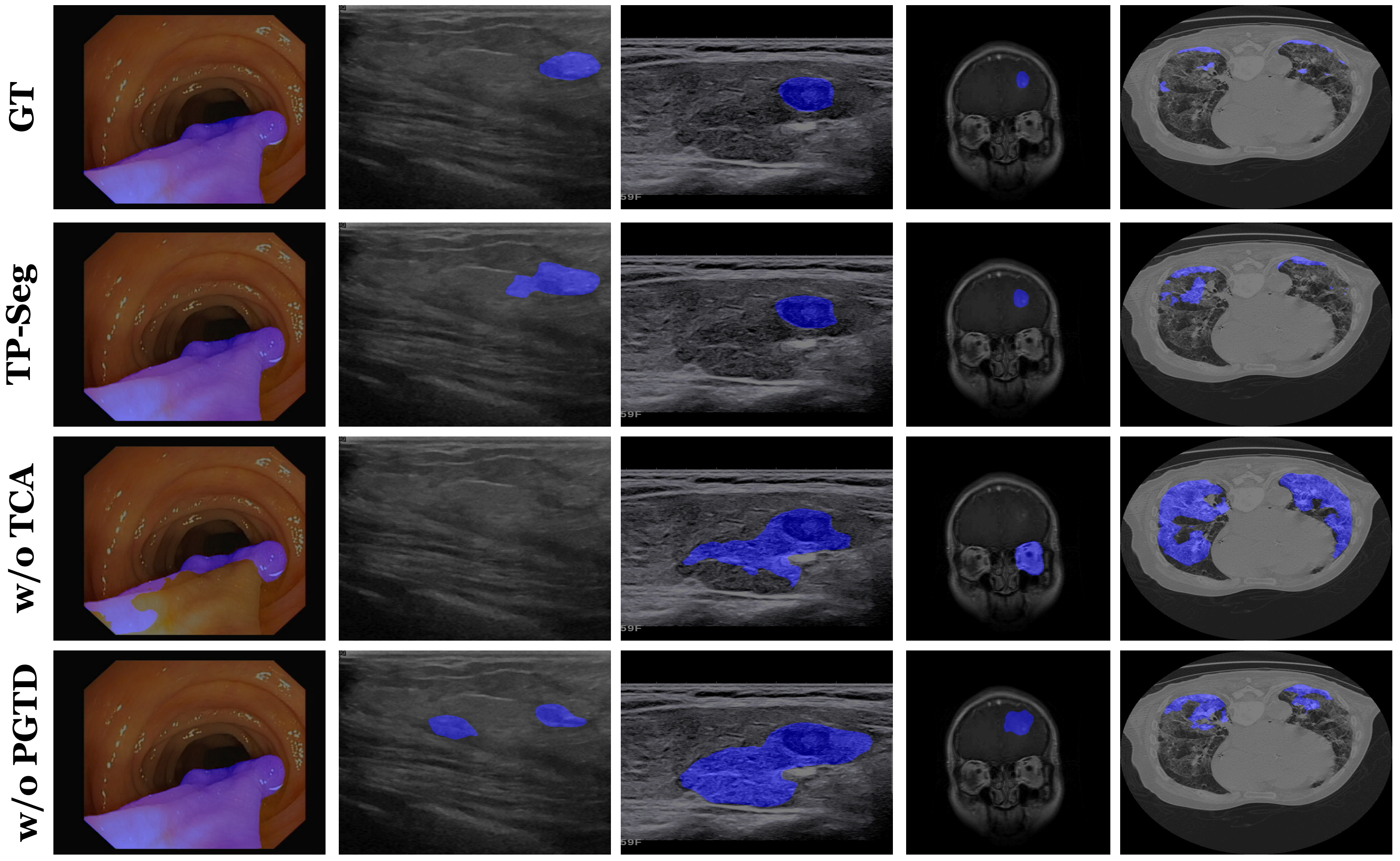}
    \caption{Visual comparison of component-wise predictions.}
    \label{fig:image6}
    \vspace{-5mm}
\end{figure}

\subsection{Qualitative Evaluation}
Fig.~\hyperref[fig:image5]{5} shows the visual comparison with three other unified models. It can be seen that TP-Seg yields predictions closest to the ground truth, accurately capturing lesions with diverse morphologies, including both large and small objects.
For large or uniquely shaped lesions such as those in WA, ADC, TN, and BL, TP-Seg exhibits superior boundary fidelity and refinement. In ADC and TN, TP-Seg provides highly smooth and spatially consistent boundaries, significantly outperforming the competing models which often show slightly fragmented predictions or extensive false positive regions. In BL, our model precisely captures the intricate delineation of the long, thin lesion, especially at its tapered ends.
Regarding small or varying-sized lesions in BT, LI, and SL, TP-Seg showcases high sensitivity and precision. For example, in SL and WA, TP-Seg successfully identifies subtle pathological areas frequently overlooked by other models. In BT and LI, it accurately isolates the target lesions while effectively suppressing background interference ensuring compact and accurate segmentation results.
For CP, TP-Seg successfully identifies and segments polyps that are deeply embedded or partially occluded which shows TP-Seg's superior sensitivity to multiple yet hidden lesions. 
Overall TP-Seg shows strong generalization and robustness when handling various medical imaging modalities and lesion characteristics.

\subsection{Ablation Study}
Tab.~\hyperref[tab:3]{3} presents our ablation results. All experimental settings remain consistent to ensure fair comparability.

\begin{figure}[t]
    \centering
    \includegraphics[width=\columnwidth]{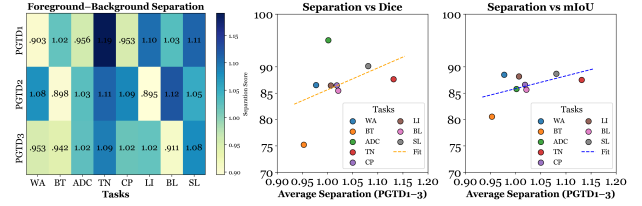}
    \caption{FG–BG separation scores across 8 MLS tasks and their correlations with segmentation performance metrics (Dice (\%) and mIoU (\%)). The separation scores range from 0 to 2. A score of 0 (cosine similarity = 1) indicates that FG and BG prototypes are nearly identical and difficult to distinguish, while a score of 2 (cosine similarity = -1) denotes perfectly opposite prototypes, representing the ideal case for clear separation.}
    % Schematic diagram illustrating the FG–BG separation scores across 8 MLS tasks, and their correlations with segmentation performance metrics (Dice (\%) and mIoU (\%)). The separation scores ranges from 0 to 2. A value of 0 (cosine similarity = 1) signifies that the FG and BG prototypes are nearly identical, making them hard to distinguish. Conversely, a value of 2 (cosine similarity = -1) indicates the prototypes are perfect opposites, which represents the ideal scenario for clear separation.
    \label{fig:image8}
    \vspace{-5mm}
\end{figure}

\noindent \textbf{Each Component.}
First, the baseline model (A1) integrates the SAM 2 encoder with a U-shaped skip-connection decoder, serving as a unified architecture without any task-specific design.
Next, when equipped with both the task-conditioned adapter (TCA) and the prototype-guided task decoder (PGTD), the full model (A4) achieves the highest performance, surpassing A1 by a large margin (+7.94\% Dice on average). This significant improvement demonstrates the effectiveness of our task-prototype framework in enhancing both feature adaptability and semantic discrimination across diverse lesion segmentation tasks.
Then, ablation variants further confirm the complementary roles of the two components. Removing TCA (A2) leads to a noticeable drop in performance due to the lack of adaptive feature modulation for different tasks and modalities, while removing PGTD (A3) results in weaker structural perception and less precise boundary localization. The consistent improvements across all 8 tasks validate the robustness and universal effectiveness of both modules. We also provide visual comparisons in Fig.~\hyperref[fig:image8]{6}, where removing either module leads to noticeably coarser predictions and blurred boundaries, clearly evidencing the crucial role of both components in achieving precise and consistent lesion segmentation.
Finally, the comparable or even superior performance of A4 against separate training (A5) highlights the strong compatibility and balance achieved by our method across heterogeneous tasks and modalities, without mutual exclusion.

\noindent \textbf{Task-Conditioned Adapter.}
First, we compare the proposed TCA with classical parameter-efficient adaptation strategies, as shown in Tab.~\hyperref[tab:3]{3 (B)}. In particular, B1 and B2 adopt LoRA and Adapter for fine-tuning, respectively, while B3 removes the task-specific route from TCA. All these variants exhibit clear performance degradation compared to the full model (A4). This decline arises because traditional lightweight adaptation methods cannot effectively handle the substantial modality gaps among heterogeneous tasks, resulting in insufficient task decoupling and impaired generalization.
Next, to further verify the necessity of TCA’s dual-route design, we remove the shared route in B4. The resulting performance drop indicates that the absence of shared cross-task representations limits inter-task knowledge transfer, causing the model to behave similarly to independently trained variants.
Finally, the consistent superiority of A4 confirms that TCA achieves an optimal balance between task independence and inter-task collaboration. By integrating both task-specific and shared feature routes, TCA effectively eliminates cross-task interference while leveraging shared information to maintain unified yet discriminative representations across diverse modalities.

% Tab.~\hyperref[tab:3]{3 (b)} presents the ablation results of TCA relative to classic parameter-efficient transfer learning methods. In B1, we employ the LoRA method for fine-tuning adaptation; in B2, we use the Adapter method for fine-tuning; and in B3, we remove the task-specific route from TCA. It can be observed that the performance of all three variants drops significantly compared to A4. This decline is mainly due to the pronounced modality differences among tasks, where these methods struggle to effectively mitigate cross-task modality conflicts, thereby weakening task independence and leading to suboptimal overall performance. To further validate the effectiveness of TCA’s design, in B4, we remove the shared route from TCA, which also results in a noticeable performance degradation compared to A4. This finding indicates that, without shared cross-task feature representations, the model fails to fully leverage inter-task information, resulting in performance similar to that of independently trained models. In summary, these experiments strongly demonstrate the superiority of TCA in unified MLS tasks, it fundamentally eliminates inter-task interference, and simultaneously leverages shared task features to achieve optimal overall performance.

\begin{table}[t]
    \centering
    \setlength{\tabcolsep}{2mm}
    \resizebox{\linewidth}{!}{%
    \begin{tabular}{c|ccccc}
    \specialrule{1.5pt}{0pt}{0pt}
    \toprule
    Average& w/o PGTD & w/o PGTD1 & w/o PGTD2 & w/o PGTD3&TP-Seg \\
    \hline
Dice $\uparrow$ &85.54& 86.16 & 86.21 & 86.41&86.63 \\
mIoU $\uparrow$ &84.85& 85.77 & 85.80 & 86.13&86.44 \\
    \specialrule{1.5pt}{0pt}{0pt}
    \end{tabular}%
    }
    \caption{Ablation study on the contribution of individual PGTD modules in terms of Dice (\%) and mIoU (\%), averaged over all 8 lesion segmentation tasks.}
    \label{tab:4}
    \vspace{-6mm}
\end{table}

\noindent \textbf{Hyperparameters.}
We further investigate the influence of three key hyperparameters in TP-Seg, as summarized in Tab.~\hyperref[tab:3]{3 (C)}.
The coefficient $\alpha$ controls the modulation strength of the prototype-guided similarity map $S_{\tau}$, which determines how effectively prototype cues refine the decoder features.
The parameter $\lambda_{p}$ regulates the contribution of the prototype-guided reinforcement term to the main prediction head $y_{\tau}$, balancing semantic consistency and adaptive flexibility across tasks.
The number of dynamic experts $M$ defines the diversity and specialization capacity of the decoder, affecting how well the model captures heterogeneous lesion patterns.
TP-Seg exhibits consistently strong performance across different configurations, with the optimal setting achieved at $\lambda_{p}=5.0$, $\alpha=0.5$, and $M=6$.
These results demonstrate the robustness of the proposed design and its resilience to parameter variations, confirming that TP-Seg maintains balanced optimization and reliable segmentation quality without excessive sensitivity to hyperparameter tuning.

\begin{figure}[t]
    \centering
    \includegraphics[width=\columnwidth]{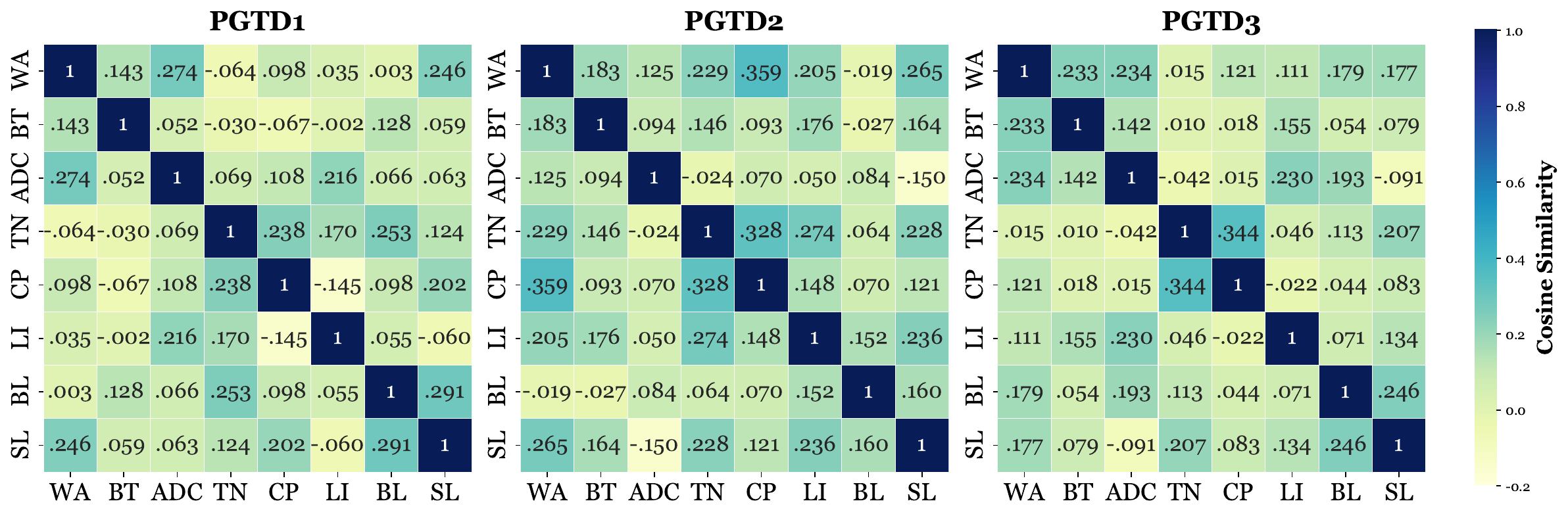}
    \caption{Cosine similarity between foreground prototypes across PGTD modules.
    }
    \label{fig:image7}
    \vspace{-6mm}
\end{figure}

\noindent \textbf{Prototype-guided Task Decoder.}
We analyze the contribution of each PGTD sub-module by removing them individually from the full TP-Seg framework. Tab.~\hyperref[tab:4]{4} reports the average results over 8 lesion tasks. Removing any module leads to noticeable performance drops, with PGTD1 having the largest impact as it generates the final prediction and integrates prototype-guided cues. The progressive decline from PGTD3 to PGTD1 highlights the hierarchical and complementary design that enhances semantic consistency and task-aware segmentation.
% We further evaluated the individual contributions of the three PGTD modules by sequentially removing each module from the original TP-Seg framework. As shown in Tab.~\hyperref[tab:4]{4}, we observe a significant performance drop with the removal of any PGTD module. Among them, PGTD1 has the most substantial impact, as it directly outputs the prediction map.

\noindent \textbf{Prototype Differentiation Analysis.}
To validate the effectiveness of the prototype-guided mechanism, we conducted a systematic visualization of prototype learning quality across all PGTD modules.
In Fig.~\hyperref[fig:image6]{7}, we compute the cosine similarity between foreground prototypes of different tasks at each PGTD layer. It can be seen that similarity across all task pairs remains well below the task collapse threshold of 0.9, consistently across all three PGTD modules. This indicates that the model successfully learns discriminative and task-specific feature representations. Notably, even semantically related tasks (e.g., Thyroid Nodule and Breast Lesion) maintain prototype similarity below 0.3, validating the effectiveness of the prototype contrastive learning strategy.
\begin{table}[t]
    \centering
    \resizebox{\linewidth}{!}{
    \begin{tabular}{lcccccc}
        \toprule
        \textbf{Module} 
        & \textbf{Params} 
        & \textbf{Growth / Task} 
        & \textbf{Growth (\%)} 
        & \textbf{FLOPs} 
        & \textbf{Growth / Task} 
        & \textbf{Growth (\%)} \\
        \midrule
        TCA
        & 3.37 M & +3.37 M & \textcolor{red}{\textbf{1.27\%}} & 1.85 G & 0.028 & \textcolor{red}{\textbf{0.01\%}} \\
        PGTD Prototypes
        & 0.46 M & +0.001 M & \textcolor{red}{\textbf{0.17\%}} & $\approx$0 & $\approx$0 & \textcolor{red}{\textbf{$\approx$0}} \\
        PGTD Decoder
        & 12.03 M & 0 & \textcolor{red}{\textbf{0}} & 10.31 G & 0 & \textcolor{red}{\textbf{0}} \\
        \midrule
        \textbf{Full Model (1 Task)}
        & 264.84 M & -- & - & 254.00 G & -- & - \\
        \textbf{Full Model (8 Tasks)}
        & 288.46 M & -- & - & 254.23 G & -- & - \\
        \bottomrule
        \specialrule{1.5pt}{0pt}{0pt}
    \end{tabular}
    }
    \caption{Comprehensive cost and scalability analysis of the proposed unified framework.}
    \label{tab:5}
    \vspace{-3mm}
\end{table}
We further evaluate the separation between foreground and background prototypes within each task. As shown in Fig.~\hyperref[fig:image7]{8}, all tasks achieve FG–BG separation scores above 0.85, demonstrating that the prototype update mechanism establishes clear and consistent foreground–background boundaries in feature space, providing strong guidance for pixel-level classification. The scatter plots further reveal a positive correlation between separation degree and segmentation performance, confirming the effectiveness of the feature separation strategy. Furthermore, Tab.~\hyperref[tab:5]{5} illustrates the parameter count of each module as the number of tasks increases, demonstrating that extending to multiple tasks incurs only a negligible increase in parameters.

% We further evaluated the separation degree between foreground and background prototypes within each task. As illustrated in Fig.~\hyperref[fig:image7]{7}, all tasks achieved FG–BG separation scores exceeding 0.85, indicating that the prototype update mechanism successfully establishes clear and consistent foreground–background boundaries in the feature space, providing strongly discriminative guidance signals for subsequent pixel-level classification. Moreover, the two scatter plots demonstrate a significant positive correlation between the separation degree and the segmentation performance metrics, further validating the effectiveness of the proposed feature separation strategy.
\section{Conclusion}
In this paper, we propose TP-Seg, a simple yet effective  framework for unified medical lesion segmentation. 
%
% We design the first task-conditioned adapter for unified medical lesion segmentation, providing a solution that not only fundamentally eliminates inter-task interference but also effectively leverages shared common features across different tasks, thereby achieving optimal overall performance. 
% %
% In addition, we introduce the Prototype-Guided Task Decoder, which utilizes continuously updated task prototypes as persistent semantic memory to achieve precise modeling of task-specific foreground and background semantics. 
%
Benefiting from the proposed task-conditioned adapter, TP-Seg effectively alleviates inter-task interference while leveraging shared representations across diverse tasks to enhance overall performance. Furthermore, with the prototype-guided task decoder that employs continuously updated task prototypes as persistent semantic memory, TP-Seg achieves precise and consistent modeling of task-specific foreground and background semantics.
Experimental results demonstrate that TP-Seg achieves strong performance across 8 medical lesion segmentation tasks, and we hope it can serve as a step toward a more generalized multi-lesion evaluation framework, offering useful insights and potential directions for future research and clinical applications.

\vspace{1mm}
\noindent \textbf{Acknowledgments.}
{\footnotesize
This work is supported by grants from the National Natural Science Foundation of China (No.62262030, No.62377011) and the Jiangxi Provincial Natural Science Foundation (No.20232BAB202021).
}

\clearpage

{
    \small
    \bibliographystyle{ieeenat_fullname}
    \bibliography{main}
}


\end{document}